\documentclass{article}

\usepackage{microtype}
\usepackage{graphicx}
\usepackage{subfigure}
\usepackage{booktabs} 
\usepackage{amsmath}

\usepackage{hyperref}

\usepackage[accepted]{icml2019}

\icmltitlerunning{2-bit Model Compression of Deep Convolutional Neural Network on ASIC Engine for Image Retrieval}

\begin{document}

\twocolumn[
\icmltitle{2-bit Model Compression of Deep Convolutional Neural Network on ASIC Engine for Image Retrieval}



\icmlsetsymbol{equal}{*}

\begin{icmlauthorlist}
    \icmlauthor{Bin Yang}{to}
    \icmlauthor{Lin Yang}{to}
    \icmlauthor{Xiaochun Li}{to}
    \icmlauthor{Wenhan Zhang}{to}
    \icmlauthor{Hua Zhou}{to}
    \icmlauthor{Yequn Zhang}{to}
    \icmlauthor{Yongxiong Ren}{to}
    \icmlauthor{Yinbo Shi}{to}
\end{icmlauthorlist}

\icmlaffiliation{to}{Gyrfalcon Technology Inc., CA, USA}

\icmlcorrespondingauthor{Bin Yang}{bin.yang@gyrfalcontech.com}

\icmlkeywords{Machine Learning, ICML}

\vskip 0.3in
]



\printAffiliationsAndNotice{}  

\begin{abstract}
Image retrieval utilizes image descriptors to retrieve the most similar images to a given query image.
Convolutional neural network (CNN) is becoming the dominant approach to extract image descriptors for image retrieval.
For low-power hardware implementation of image retrieval,
the drawback of CNN-based feature descriptor is that it requires hundreds of megabytes of storage.
To address this problem, this paper applies deep model quantization and compression to CNN in ASIC chip for image retrieval.
It is demonstrated that the CNN-based features descriptor can be extracted using as few as 2-bit
weights quantization to deliver a similar performance as floating-point model for image retrieval.
In addition, to implement CNN in ASIC,
especially for large scale images, the limited buffer size of chips should be considered.
To retrieve large scale images, we propose an improved pooling strategy,
region nested invariance pooling (RNIP), which uses cropped sub-images for CNN.
Testing results on chip show that 
integrating RNIP with the proposed 2-bit CNN model compression approach is capable of retrieving large scale images. 
\end{abstract}

\section{Introduction}
\label{submission}
Image retrieval employs compact feature descriptors to locate target images
containing the same object or scene as the one depicted in a query image \cite{iamge_retrieval_1,iamge_retrieval_2,iamge_retrieval_3}.
With the rapid progress of deep neural networks over the last few years,
convolutional neural network (CNN) is becoming the dominant approach
and has been adopted by MPEG for generating descriptors \cite{ISO/IEC_1,hnip,mpeg}.

\par 
State-of-the art CNN networks consist of hundreds of millions of neurons,
which require hundreds of megabytes for storage if being represented in floating-point precision \cite{model_compression_1}.
Image retrieval applications with CNN would require
descriptor extraction to be performed with limited hardware resources.
Therefore, it is preferred to use fixed point format with fewer bits to reduce both logic and memory space
and accelerate the CNN processing.
Meanwhile, CNN model compression is broadly used in hardware with fixed-point format
to reduce the computation cost and the model size stored in local memory \cite{model_compression_2}.
The model compression methods include pruning, coefficients clustering and quantization, etc. \cite{xnor,model_compression_3}.
The scalar and vector quantization techniques and a scheme that shares weights across layers
have been used to reduce the CNN model size, and negligible loss is achieved in retrieval performance
with 4 bits quantization for VGG-16 \cite{instance}.
It has also been demonstrated that the CNN-based descriptor can be extracted using 4 bits compressed CNN model
on devices with memory and power constraints \cite{ISO/IEC_2}.
In addition, the limited image buffer size of chips should be considered for implementing CNN in ASIC,
especially for large scale images.

\par In this paper, we propose a deep model compression and weight quantization approach to enable efficient hardware implementation
for CNN-based descriptor extraction.
Our compression method balances compressed bit precision and performance using a hybrid bit quantization scheme across layers.
The compressed model with as few as 2-bit weights quantization
can deliver a similar performance as the floating-point model for image retrieval.
Meanwhile, to handle large scale images,
we use a region nested invariance pooling (RNIP) strategy
based on the concatenation of feature maps from different sub-images
to extract deep feature descriptor.
This proposed RNIP approach is compatible with and can be combined with the deep model compression technique
to enable the large scale image retrieval.

\par The original contributions of this paper are as follows:
\vspace{-7pt}
\begin{itemize}
\setlength{\itemsep}{-1pt}
\item Using as few as 2-bit weights quantization of CNN on ASIC-chip,
we achieve a similar image retrieval performance as the floating-point model;
\item We propose an improved pooling strategy, RNIP, that can be integrated with the 2-bit CNN model compression approach
to retrieve large scale images.
\end{itemize}
\vspace{-7pt}

\par The paper is organized as follows.
Section 2 shows how CNN and NIP are used for image retrieval. 
The proposed compressed model is presented in Section 3.
The proposed RNIP strategy is described in Section 4. 
Section 5 shows the results of the compressed model and RNIP on chip for image retrieval.
Section 6 concludes this paper.

\section{Image retrieval with CNN}
For image retrieval, CNN and NIP is standardized as a reference model to generate deep feature descriptor \cite{ISO/IEC_1}. 
In the first step, multiple rotated images are created from the input image at 0, 90, 180 and 270 angles.
Then, each rotated image is fed into CNN to generate convolutional feature maps. Following that, 
NIP is used to generate a single deep feature descriptor from feature maps of each rotated image by utilizing
square-root pooling, average pooling, and max pooling in a chain \cite{hnip,mpeg,nip}.
At last, a 512-D vector is generated with VGG-16 model.

\section{Deep Model Compression}
\subsection{Quantization}

In a CNN, each convolutional layer is made up of neurons that consist of learnable weights and biases.
Consider the $k$th weight and the $k$th bias of each layer defined by ${W_{k}}$ and ${b_{k}}$.
For each layer, our proposed quantization and compression scheme is as follows:

\par 1) The bias ${b_k}$ is represented as a 12-bit signed integer.
\par 2) To quantize ${W_{k}}$, the $i$th 3x3 filter {${W}_{ki}$} can be expressed as
\begin{equation}
    W_{ki}\approx\alpha_{ki}{M}_{ki}
\end{equation}
where $\alpha_{ki}$ is the scalar for the $i$th filter and is quantized to 8 bits;
${M}_{ki}$ is the quantized 3x3 mask for the $i$th filter which can be quantized to 1, 2, 3, 4 or 5 bits for different layers.
Consider the $l$th element of ${M}_{ki}$ and the $l$th element of ${W}_{ki}$ defined by $m_l$, $w_l$, respectively.
In the proposed quantization and compression scheme, for 1-bit quantization,
$m_l$ is projected into 1 if $w_l>\alpha_{ki}$, otherwise -1, with $\alpha_{ki}$ being a threshold,
where the scalar $\alpha_{ki}$ is calculated by the mean of ${W}_{ki}$.

\par3) To accommodate the dynamic range of the filter coefficients, a layer-wise 4-bit shifting value represents the exponent value of the weights.


    \begin{algorithm}[ht]
        \caption{re-train fixed point model}
        \label{alg:example}
     \begin{algorithmic}
        \STATE {\bfseries Input:} Weights $[W_1, W_2, ..., W_K]$ for each layer
        \STATE Find the maximum value $max value$ of Weights
        \FOR{$k=1$ {\bfseries to} $K$}
        \FOR{$i=1$ {\bfseries to} $I$}
        \STATE Quantize $\alpha_{ki}$ and $M_{ki}$ using $max value$
        \STATE Update $W_{ki}$ using Equation (1)        
        \ENDFOR
        \ENDFOR
        \STATE Do backpropagation 
     \end{algorithmic}
     \end{algorithm}
\par To quantize the floating-point coefficients, it is important to trade-off its range and precision.
Directly quantizing the model weight parameters will affect the model accuracy.
It is discovered that re-training the model with the constraints of fixed point weights would make 
the model's precision very close to the floating point model's precision with far fewer bits used for model weights.
Algorithm 1 is used to re-train the fixed-point model.

\subsection{Weights Precision and Compression Ratio}

Regarding the 32-bit floating point model, the compression ratio {$r$} can be calculated as $r=(32*{z})/({z}*{m}+{s})$,
where $z$ is set as 9 for 3x3 filter, $m$ is the number of bits for each element in mask, and $s$ is the number of bits for scalar.
In this work, we quantize the coefficients with different precision for different layers, where
the earlier layers use more bits for the mask to improve accuracy without increasing the model size too much.
For the first seven convolution layers of VGG-16, we use 3 bits for masks, resulting in a 8.2 times compression ratio.
The last six convolution layers use 1-bit mask; it is about a 17 times compression ratio compared to the 32-bit floating point model.
For VGG-16, this achieves an overall compression ratio of about 15.1 times, which is close to 2-bit fixed point quantization.
The proposed compressed model can be easily implemented in ASIC hardware design for image retrieval.

\section{Region Nested Invariance Pooling}
\begin{figure}[ht]
    \vskip -0.01in
    \begin{center}
    \centerline{\includegraphics[width=\columnwidth]{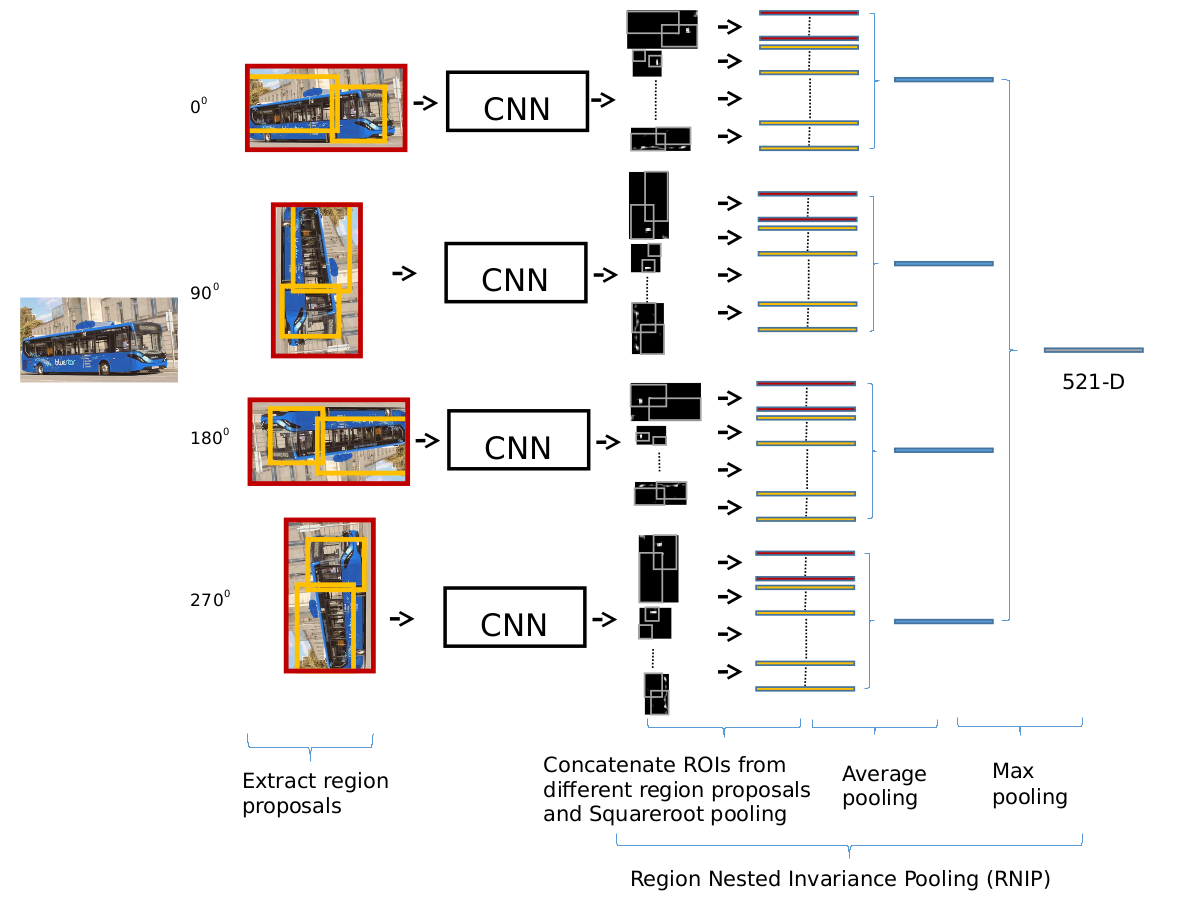}}
    \vspace{-7pt}
    \caption{Region Nested Invariance Pooling}
    \label{Pyramid Nested Invariance Pooling}
    \end{center}
    \end{figure}  
With reference to object detection and image segmentation using cropped sub-images,
this paper proposes an improved NIP strategy, called RNIP, for deep feature descriptor extraction.
As shown in Fig. 1, the RNIP strategy uses cropped sub-images from the original image as the input of CNN.
The sub-images as CNN input can reduce the buffer size on devices.
First, we crop the input image to generate multiple size sub-images,
using the same crop strategy as region of interesting (ROI) samplings of feature maps in standard NIP.
Second, each cropped image is fed into CNN to generate convolutional feature maps.
Following that a set of ROIs from feature maps of corresponding cropped images are concatenated.
Finally, NIP pooling is performed on the set of ROIs to generate a single descriptor.
For VGG-16, RNIP generates a 512-D feature descriptor for image retrieval.

\section{Experiment Results}
\subsection{Image Retrieval with RNIP and NIP}
In this section, examples are presented for the performance of RNIP on image retrieval as shown in Table 1.
We build descriptors with floating-point VGG-16.
The image retrieval performance is evaluated by mean average precision (mAP).

In Table 1, we resize all images to 224x224 on INRIA Holidays dataset for image retrieval. 
RNIP 5x and RNIP 14x use 5 and 14 cropped sub-images to extract deep feature descriptors, respectively;
512-D, 512-byte and 512-bit mean that each descriptor element is single-precision floating point,
8 bits quantized and binarized descriptors,
respectively.

\begin{table}[ht]
    \vskip -0.15in
    \caption{RNIP vs. NIP for Image Retrieval}
    \label{table1}
    \begin{center}
        \begin{small}
            \begin{tabular}{lccr}
                \toprule
                \begin{tabular}{@{}c@{}}Descriptor \\ size\end{tabular} &
                    NIP &
                \begin{tabular}{@{}c@{}}RNIP \\ 5x\end{tabular}&
                \begin{tabular}{@{}c@{}}RNIP \\ 14x\end{tabular}\\    

                \midrule
                512-D          & 0.806     & 0.842 & 0.852   \\
                512-byte     & 0.806     & 0.841 & 0.852   \\
                512-bit      & 0.773     & 0.803 & 0.824   \\
  
                \bottomrule
            \end{tabular}            
        \end{small}        
    \end{center}
    \vskip -0.1in        
\end{table}


  

\par It clearly shows that image retrieval results for 512-byte descriptor and 512-D single-precision floating descriptor are very close.
Hence, the descriptor size per image can be reduced from 2 KB to 0.5 KB without sacrificing image retrieval performance.
For binaried descriptor, it is obvious that image retrieval performance drops around 0.03 mAP.
In Table 1, it is clearly seen that the RNIP can improve the performance of image retrieval.

\subsection{Image Retrieval using ASIC Engine}
For classification task, the proposed retrained fixed-point VGG-16 model has achieved T1:70.13{\%} and T5:89.91{\%} accuracy on ILSVRC-2012 dataset,
using as few as 2-bit weights quantization of VGG-16 on ASIC chip.

The performance of image retrieval using ASIC engine and NIP on INRIA Holidays dataset is shown in Fig. 2.
It shows that the retrained fixed-point VGG-16 model using 2-bit weights quantization achieves the same performance as the floating-point model for
both 224x224 and 640x480 input image size.
In Fig. 2 (a), it is clearly shown that our proposed deep model compression and quantization scheme on CNN ASIC Engine
delivers the same image retrieval performance as the floating-point model.

   
  

\begin{figure}[t]
    \begin{center}
    \centerline{\includegraphics[width=85mm]{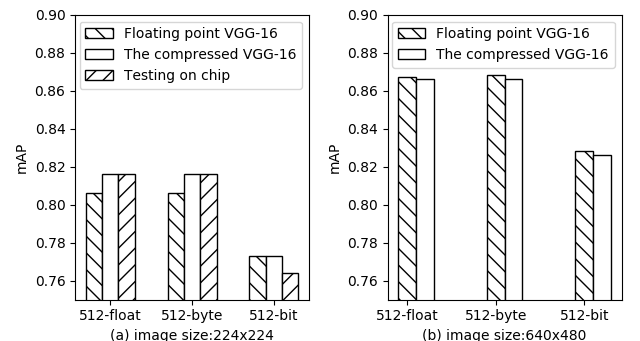}}
    \vspace{-7pt}
    \caption{Image Retrieval Hardware Implementation}
    \label{Image Retrieval Hardware Implementation}
    \end{center}
    \vskip -0.2in
    \end{figure}
\par In Table 2, it clearly shows that the proposed RNIP can extract deep features locally from sub-images
with the same image retrieval performance.
The proposed model compression scheme reduces the VGG-16 model size from 59M bytes to 4M bytes.
It clearly shows that, using the proposed RNIP, 
VGG-16 using the proposed 2-bit weights quantization approach on chip with 224x224 image input 
can deliver the same performance of image retrieval as 640x480 image input for VGG-16.

\begin{table}[ht]
    \vskip -0.15in
    \caption{Large Scale Image Retrieval using RNIP}
    \label{sample-table}
    \begin{center}
    \begin{small}
    \begin{tabular}{lccr}
    \toprule
                             & VGG16+NIP & ASIC+RNIP  \\
    \midrule
    Input size:              & 640x480   & 640x480 \\
    Input cropping:          & N/A       & 9x crop\\
    CNN model input size:    & 640x480   & 224x224 \\
    CNN model format:        & Floating  & 2-bit fixed-point  \\
    CNN model size:          & 59M bytes & 4M bytes\\
    Descriptor size:         & 512 bytes & 512 bytes \\
    mAP score:               & 86.81{\%} & 86.94{\%}   \\
 
    \bottomrule
    \end{tabular}
    \end{small}
    \end{center}
    \vskip -0.1in
    \end{table}

\section{Conclusion}
We proposed a deep model compression and quantization scheme as well as an improved pooling strategy named RNIP
to enable efficient hardware implementation of CNN-based descriptor for image retrieval.
Experimental results show that our compressed VGG-16 with as few as 2-bit quantization
can be directly ported onto our specially designed ASIC CNN engine,
while achieving a similar performance to the floating-point model.
Integrated with RNIP strategy, the model with the proposed 2-bit quantization approach can retrieve 640x480 images
with a limited 224x224 image buffer on the ASIC chip with similar performance.

\nocite{langley00}

\bibliography{example_paper}
\bibliographystyle{icml2019}



\end{document}